\documentclass[10pt,twocolumn,letterpaper]{article}
\usepackage[T1]{fontenc}
\usepackage{lmodern}
\usepackage[margin=0.75in,columnsep=0.25in]{geometry}
\usepackage{amsmath,amssymb,booktabs,graphicx,microtype,array}
\usepackage{xcolor,tikz}
\usetikzlibrary{positioning,arrows.meta}
\usepackage[hidelinks]{hyperref}
\usepackage{url}
\usepackage{enumitem,placeins}
\setlist{leftmargin=*,itemsep=2pt,topsep=3pt}
\newcommand{\I}{\mathcal{I}}
\newcommand{\N}{\mathcal{N}}
\newcommand{\model}{VINTAGE-TS}
\title{\vspace{-1em}\textbf{Time-Series Foundation Models\\That Understand Data Revisions}}
\author{Taimoor Ahmed \thanks{Superior University Lahore, Pakistan}}

\hypersetup{pdftitle={Time-Series Foundation Models That Understand Data Revisions}
}

\begin{document}
\raggedbottom
\maketitle
\begin{abstract}
Historical observations are not always fixed: statistical agencies revise previously published values as new evidence arrives. Forecasting from a contemporary download can therefore expose a model to information unavailable at the date it purportedly made a prediction. We propose \model{}, a revision-aware adaptation of a time-series foundation model that distinguishes observation time from information-availability time. Its targets are the next period's first-published value and the value available a fixed number of days after that publication; neither is declared final truth. A joint predictive distribution preserves dependence between these targets and exposes uncertainty about their difference. We specify an ALFRED-based rolling evaluation, a matched Chronos-2 comparison, conventional and revision-aware baselines, and a separate audit of pretraining overlap. The accompanying software implements validity-interval reconstruction, delayed-label filtering, a frozen-backbone adapter interface, and reproducible diagnostics. An executed synthetic demonstration and a 25-configuration sensitivity suite verify the workflow, expose variation across seeds and revision regimes, and illustrate how hindsight contamination changes measured performance. Thirty-one automated tests check temporal and integration contracts. Real ALFRED and Chronos-2 experiments have not been executed; no empirical foundation-model advantage is claimed.
\end{abstract}


\section{Introduction}
A forecaster operates on the information available at the time a decision is made. Many forecasting benchmarks implicitly identify that information with the historical part of a series downloaded today. This identification fails when old observations are revised. An employment estimate for January can have one value in February, another in March, and a substantially different value after an annual benchmark revision. All describe the same observation period, but they become usable at different dates \cite{croushore2001,croushore2011}.

ALFRED preserves historical vintages and supports reconstruction of previously available data \cite{alfred}. This makes revisions an observable forecasting problem rather than a synthetic perturbation of an otherwise fixed series. Two questions then arise: what will the next release report, and what will that reported observation become after a specified maturation interval? These questions serve different users and need not favor the same model; forecast assessment can depend on which vintage defines the outcome \cite{stark2002}.

Chronos-2 provides a pretrained starting point with multivariate and covariate-informed forecasting support \cite{chronos2}. Those capabilities make it feasible to supply release-age and revision features alongside a current vintage. They do not, on their own, define the target vintage, enforce historical availability, or prove that pretraining avoided later revisions. The proposed contribution is the integration of explicit information timing, separate release targets, and a joint uncertainty model with that backbone.

This manuscript provides three concrete components. First, it defines an operational data and target contract with no assumption of a final value. Second, it specifies a revision-aware frozen-backbone adaptation and an extension to joint neural training. Third, it supplies an executable reference package with rolling evaluation, information audits, and a synthetic smoke experiment. The empirical research claim remains a hypothesis: access to revision histories may improve mature-vintage forecasts or their calibration beyond a matched latest-vintage model. The protocol is designed to reject that hypothesis if the evidence does not support it.

\section{Positioning and Research Questions}
Real-time forecasting and data revisions have a substantial econometric history. Croushore and Stark construct a real-time macroeconomic data set \cite{croushore2001}, and Stark and Croushore examine forecasting with it \cite{stark2002}. Koenig, Dolmas, and Piger show that the vintage strategy used to estimate a forecasting equation can affect its out-of-sample performance \cite{koenig2003}. Aruoba documents departures from commonly imposed assumptions about revision behavior \cite{aruoba2008}, while Croushore surveys the wider real-time data literature \cite{croushore2011}. Jacobs and van Norden develop state-space representations of revisions that accommodate richer measurement-error dynamics and do not require an observed vintage to equal an ultimate true value \cite{jacobs2011}. Consequently, two time coordinates and revision modeling alone are not novel contributions.

A related literature studies information arriving at different publication dates. Giannone, Reichlin, and Small develop a framework for real-time macroeconomic nowcasting \cite{giannone2008,a12}. Ba\'nbura and Modugno address factor-model estimation with arbitrary missing-data patterns, including those associated with different release delays \cite{banbura2014,a11}. These contributions motivate attention to asynchronous availability. Here, the additional object of interest is the changing sequence of published values for the same observation period.

Neural forecasting provides several relevant modeling ingredients. DeepAR learns probabilistic forecasts across related series \cite{salinas2020,a10}; Temporal Fusion Transformers combine multiple input types for multi-horizon prediction \cite{lim2021,a9,a92}; and PatchTST studies patch-based Transformer representations \cite{nie2023,a8}. Published work on Chronos, TimesFM, and Moirai further establishes pretrained and universal forecasting as an active research direction \cite{chronos2024,das2024,woo2024,a7}. These papers motivate the backbone and representation choices; they do not establish that our proposed revision-aware adaptation improves forecasting.

The evaluation also draws on established methodology. Proper scoring rules assess predictive distributions \cite{gneiting2007scores,a6}, and calibration must be considered together with sharpness \cite{gneiting2007calibration,a5}. Rolling-origin testing and predictive-accuracy comparison have a substantial forecasting literature \cite{tashman2000,diebold1995,a4}. Validation assumptions matter for dependent observations: results permitting ordinary cross-validation for some autoregressive settings do not by themselves authorize access to future vintages \cite{bergmeir2018,a3}. More broadly, leakage can make machine-learning results appear stronger than they are \cite{kapoor2023,a2}. This motivates an explicit information audit; it is not evidence that any particular Chronos checkpoint has seen the evaluation targets.

The intended research gap is narrower: how should a broadly pretrained forecasting model consume revision history, learn from labels that mature at different dates, and produce forecasts for explicitly named vintages under a defensible information budget? This manuscript does not establish a first-in-literature claim. A broader systematic review remains necessary before a submission claims novelty over all revision-aware neural models.

We separate three research questions:
\begin{enumerate}
\item Does revision history add predictive value beyond the latest vintage available at issuance, holding backbone, target definition, and training budget fixed?
\item Can a joint forecast distinguish uncertainty about the first release from uncertainty about its subsequent revision, while maintaining useful marginal calibration?
\item How much do data-vintage leakage and uncertain pretraining exposure change the interpretation of apparent forecasting gains?
\end{enumerate}
Success requires evidence on the first two questions after the third has been audited. Better performance under a hindsight-contaminated input is not evidence of a deployable improvement.

\section{Two-Time Data and Target Contract}
\subsection{Availability and validity intervals}
Let $i$ index a series and $t$ an observation period. Let $v$ denote an information-availability date. An archived record is
\begin{equation}
 e=(i,t,v_e,u_e,x_e),
\end{equation}
where $[v_e,u_e]$ is the closed interval during which the recorded value is current, following the API's inclusive real-time-period semantics \cite{fredapi,realtime}. At end-of-day forecast origin $o$, the current value is
\begin{equation}
 y_{i,t}^{[o]}=x_e \quad\text{if }v_e\leq o\leq u_e.
\end{equation}
The historical event set contains records with $v_e\leq o$. Their eventual end dates are used only to resolve interval membership; a future end date is never an input feature. Revision counts, first values, and differences are computed from the truncated event history.

The information set $\I_o$ includes those events and metadata demonstrably available by $o$. A forecast is a function of $\I_o$, trained using eligible earlier labels. Different models may use different summaries of the same information set, but no model receives records unavailable to another at that date. A deliberately contaminated diagnostic is an explicitly marked exception.

The reference runner uses monthly period-start timestamps and issues a forecast on the day before the target month begins. Thus, it forecasts the next observation period's eventual first publication; it does not use a future realized release calendar to select an origin. The forecast horizon from the last available observation can exceed one month because publication is delayed. Intraday forecasts, quarterly series, and mixed-frequency reconciliation require extensions.

\subsection{First and mature targets}
Let $r_{i,t}$ be the actual first-publication date, when verified by the archive-coverage audit. For a prespecified maturity lag $L$ in calendar days, define
\begin{align}
Y^{F}_{i,t}&=y_{i,t}^{[r_{i,t}]},\\
Y^{M,L}_{i,t}&=y_{i,t}^{[r_{i,t}+L]},\\
D^{L}_{i,t}&=Y^{M,L}_{i,t}-Y^{F}_{i,t}.
\end{align}
The mature target is the value \emph{in force} at the fixed date $r_{i,t}+L$. It is not the next revision after that date, the $k$th release, or the latest value at download. If no revision occurs by that date, the preceding still-valid value remains the target. If the archive does not cover the date, or the value is missing then, the mature label is unavailable. The default is $L=180$ days; $90$- and $365$-day sensitivities are included in the synthetic suite and remain to be evaluated on audited real data.

The earliest archived record need not be the original first publication. A historical series may enter the archive after many revisions. The software therefore calls its raw label ``first archived'' internally and requires a series-specific coverage audit before a real-data run is interpreted as first-release forecasting. Requesting the earliest possible API real-time date does not establish that coverage.

\subsection{Label eligibility and transformations}
At training origin $o$, a first label is eligible only if $r_{i,t}\leq o$; a mature label is eligible only if $r_{i,t}+L\leq o$. The reference adapter uses complete target pairs, applying the second and stricter condition. This simplifies a fair joint comparison but wastes some recent first-release labels. A neural implementation should use target-specific loss masks to exploit them safely.

Following the distinction between real-time and retrospectively revised estimation data \cite{koenig2003,a1}, every historical training feature vector is reconstructed at its own historical origin. Its target may later become available for training, but its context is never retrospectively upgraded. Feature centering and scaling for the fitted head use only its training partition. Per-origin level and scale features use only that origin's available history. Forward filling is causal and retains an explicit missingness channel; backward filling is excluded.

For economic applications, raw levels, log changes, or growth rates must be chosen before the test. A change computed from a vintage must use both operands available in that vintage. The executable reference uses levels and performs no seasonal adjustment or rebasing. Unit changes, definition changes, and benchmark breaks must be recorded and may require cohort separation.

\section{Revision-Aware Adaptation}
\begin{figure*}[t]
    \centering
    \includegraphics[width=\textwidth]
        {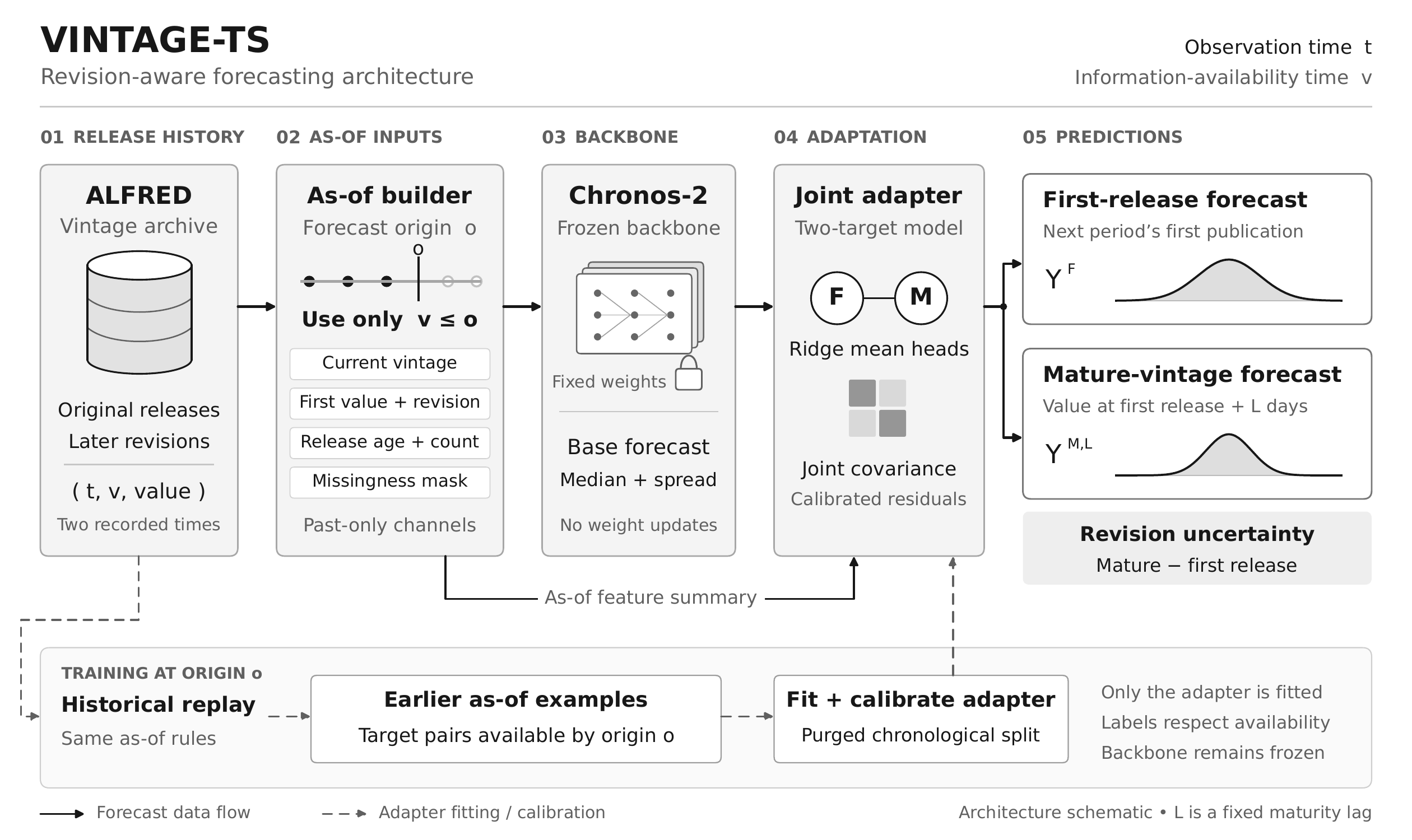}
    \caption{VINTAGE-TS architecture. Historical releases are
    reconstructed using only information available at forecast
    origin \(o\). A frozen Chronos-2 backbone and a calibrated
    joint adapter produce first-release and mature-vintage
    forecasts, together with revision uncertainty. Adapter
    fitting uses historical as-of examples and labels available
    by the corresponding cutoff.}
    \label{fig:system}
\end{figure*}

\subsection{Input channels}
For each visible observation period, construct the latest value, first archived value, cumulative revision, publication age, event count, and missingness indicator. All are functions of $\I_o$. A sparse revision triangle could retain every event; the reference implementation compresses it into these channels. This is a deliberate approximation: two distinct revision paths with identical summaries are indistinguishable to the adapter.

The latest-vintage comparator receives only the current-value sequence. The revision-aware mode also passes the additional channels as \emph{past-only covariates}. Future revision sizes, future event counts, and realized publication dates are never passed as known future covariates. The public Chronos-2 dataframe interface distinguishes historical covariates from supplied future covariates \cite{chronoscode}. The same checkpoint is used in both modes, and inference is independent across series in the reference runner.

\subsection{Executable frozen-backbone adapter}
Let $b_o$ be the frozen backbone's median for the target observation period. Let $s_o>0$ be a scale derived from differences in the as-of context, and $z_o$ the feature vector. A two-output residual adapter models
\begin{equation}
\begin{bmatrix}Y^F\\Y^{M,L}\end{bmatrix}\Bigm|\I_o
\sim \N\!\left(
 b_o\mathbf{1}+s_o B^\top\widetilde z_o,
 s_o^2\widehat\Sigma_o\right).
\label{eq:adapter}
\end{equation}
Here $\widetilde z_o$ contains an intercept and training-standardized features. The reference head is ridge regression with penalty 10 and an unpenalized intercept. It uses 12 context lags, the missingness indicators, and, in revision-aware mode, the corresponding revision, log-age, and log-count features. Chronos modes additionally include the normalized backbone median and an interval-derived spread feature. The Gaussian distribution is an adapter approximation; it is not asserted to be the native Chronos predictive law.

Both Chronos modes use the same head type, training rules, and calibration budget. This produces a controlled comparison between \texttt{Chronos-latest} and \texttt{VINTAGE-TS-frozen}. The latter is an adaptation of pretrained weights, not end-to-end foundation-model training. Without a checkpoint, the CPU-only linear reference sets $b_o$ to the latest available value and compares corresponding feature sets.

The residual covariance is estimated from the most recent 12 complete, eligible calibration pairs. The mean head is trained only on pairs whose mature target dates precede or equal the first calibration origin. This purges overlap caused by delayed labels. Calibration pairs are disjoint from mean-fitting pairs, and the head is not refitted on them afterward. A diagonal shrinkage term and a small ridge ensure positive covariance eigenvalues. The procedure estimates local empirical uncertainty; it does not establish finite-sample coverage under arbitrary distribution shift.

The two-target covariance yields a revision distribution directly:
\begin{align}
\mu_D&=\mu_M-\mu_F,\\
\sigma_D^2&=\Sigma_{MM}+\Sigma_{FF}-2\Sigma_{FM}.
\end{align}
This is preferable to adding marginal variances under an unjustified independence assumption. More generally,
\begin{equation}
\operatorname{Var}(Y^M)=\operatorname{Var}(Y^F)+\operatorname{Var}(D)
+2\operatorname{Cov}(Y^F,D).
\end{equation}
These are forecast-distribution identities, not a causal identification of measurement error or an assertion that revisions are independent news.

\subsection{Proposed neural extension}
A subsequent model can encode each event with separate observation-time and release-time embeddings and attend only to events visible at its example origin. Two heads, or a joint conditional density, would predict $Y^F$ and $D^L$. The joint construction would recover $Y^M=Y^F+D^L$. This supports revision nowcasting for already-published periods as well as forecasting future releases.

For marginal quantile heads, an eligible training objective is
\begin{equation}
\mathcal L=\sum_{k\in\{F,M\}}\lambda_k
\sum_{e,q}m_{e,k}(o)\rho_q(Y_{e,k}-\widehat Q_{e,k,q}),
\end{equation}
where $m_{e,k}(o)$ enforces target-specific availability. A joint likelihood or a dependence model is additionally needed to obtain a meaningful revision distribution; two collections of marginal quantiles alone do not identify it. The event encoder, neural training loop, mixed-frequency batching, and partially observed joint loss are \emph{proposed extensions}, not implemented or tested components of this package.

\section{Evaluation Protocol}
\subsection{Archive construction}
Download ALFRED validity intervals through the FRED observations endpoint with explicit real-time bounds, identity units, and pagination \cite{fredapi,realtime}. Preserve raw responses, checksums, series IDs, and the archive snapshot date. A request with no explicit historical window is insufficient for an as-of study. The software keeps closed validity intervals and rejects overlapping duplicates instead of silently selecting a row.

ALFRED's documentation distinguishes release dates from ingestion and notes that date provenance can depend on the source or data provider \cite{alfred}. The proposed evaluation therefore uses archive-date end-of-day semantics. Source-calendar matching and a one-business-day conservative delay are useful sensitivity checks, particularly for same-day decisions. The latter sensitivity is not included in the initial runner.

Start with audited monthly employment and industrial-production series, then expand only when vintage coverage and transformations have been checked. The supplied coverage file is intentionally marked pending; it cannot authorize a first-release interpretation by default. Series selection should reflect revision regimes and availability, not test-set gains. Quarterly national accounts form a separate extension with their own release calendar and frequency handling.

\subsection{Rolling origins and comparator fairness}
At each origin, reconstruct the information set, retain eligible training examples, fit or update permitted parameters, issue immutable forecasts, and score them only after the target-vintage date. The reference program retrospectively reconstructs this process; it does not operate a live forecast registry. A prospective study should hash and timestamp predictions before release.

Use a chronological development period to select lags, regularization, feature sets, and maturity sensitivity. Freeze those choices before the final rolling period. Evaluation origins remain identical across methods, and paired comparisons use their shared target support. Report excluded origins, missing targets, and effective sample size. Mature targets close to the archive snapshot are right-censored, not filled with a latest downloaded value.

\begin{table}[t]
\centering
\caption{Comparators and implementation scope. ``As-of'' always refers to the forecast origin.}
\label{tab:baselines}
\small
\begin{tabular}{@{}>{\raggedright\arraybackslash}p{0.35\columnwidth}>{\raggedright\arraybackslash}p{0.58\columnwidth}@{}}
\toprule
Method & Input and role\\
\midrule
AR & As-of values; regularized linear autoregression with two target heads.\\
Revision-linear & Same linear family with revision-age features; backbone-free ablation.\\
Revision-state-space & As-of release events; latent level and decaying release error.\\
Chronos-latest & Frozen checkpoint with current vintage only; calibrated residual head.\\
VINTAGE-TS-frozen & Same checkpoint and head budget; revision covariates added.\\
AR-hindsight & Later revisions substituted into already-visible periods; invalid-context diagnostic.\\
\bottomrule
\end{tabular}
\end{table}

The state-space reference uses a random-walk level and a per-period release error:
\begin{align}
x_t&=x_{t-1}+\eta_t,\\
y_t^{[r_t+a]}&=x_t+\rho^{a/30}b_t+\epsilon_{t,a}.
\end{align}
The shared $b_t$ induces dependence across releases of one observation. Each archived change event is assimilated once; unchanged daily snapshots are not repeatedly treated as independent measurements. The code uses a 72-period context, $\rho=0.65$, and empirical scale defaults estimated from as-of data. It is a transparent baseline, not a replication of Jacobs and van Norden or an optimized news-and-noise model. A publication study should additionally fit richer state-space specifications with the same pretest tuning discipline.

\subsection{Accuracy, calibration, and dependence}
Report MAE and RMSE for both targets, together with CRPS, quantile loss, 80\% interval coverage, width, and probability-integral-transform diagnostics. The reference code uses analytic Gaussian CRPS and a seven-quantile grid. PIT uniformity and interval coverage are diagnostics, not established properties of the model. Also inspect the revision difference, whose covariance and predictive scale are exported per forecast.

Raw errors across differently scaled economic series should not be pooled as the primary real-data metric. Report series-level metrics and normalize by a development-period naive forecast scale before macro-averaging. The current runner exports series identifiers and all predictions but its convenience summary pools raw errors. The synthetic series share units; the paper's demonstration uses that summary. A final ALFRED analysis must implement the prespecified scale normalization.

Use paired differences on shared dates and block-bootstrap uncertainty, retaining series observed at the same origin in the same date block. The reference script implements a circular moving-block bootstrap with 1,000 draws and six-month blocks. Its intervals are descriptive for the small demonstration. The real study should report sensitivity to block length, cohort-level uncertainty, and multiplicity correction for multiple primary claims. Failure to reject equal performance should not be called equivalence.

\subsection{Decisive plots and ablations}
A revision heatmap shows how each observation changes across archived versions. It must distinguish unseen entries from zeros. Error versus publication lag should compare predictions of the same mature target after different amounts of revision history have arrived. The shipped lag plot is a carry-forward nowcast diagnostic; it is explicitly not a revision-aware neural forecast result.

The hindsight diagnostic replaces values for periods already visible at the historical origin with their archive-snapshot values. It retains the historical period support, issuance schedule, and label-eligibility rules. This isolates one common contamination path. It does not measure the full effect of also leaking future periods, using future normalization, or training on unavailable labels.

The primary ablation removes revision covariates while keeping the same Chronos checkpoint and adapter. Further proposed ablations remove age, first-vintage, or event-count channels; vary maturity lag; compare frozen and fine-tuned weights; and stratify high- and low-revision regimes. They must be selected using development data. The basic linear feature and honest/hindsight comparisons have been executed, together with the synthetic seed, maturity-lag, and revision-amplitude sensitivities in Section~\ref{sec:extended}. The matched Chronos comparison remains unrun.

\section{Pretraining Exposure Audit}
An as-of dataframe prevents input leakage but cannot erase information already encoded in pretrained weights. A model downloaded today may have been exposed to observations or revisions published after a historical forecast origin. This risk is separate from supervised fine-tuning leakage. The Chronos-2 report describes its training data and includes an inventory of real univariate corpora \cite{chronos2}; dataset names alone are not a per-series overlap certificate.

For every checkpoint, record the exact immutable revision or local weight hash, acquisition date, known training cutoff, corpus evidence, and overlap checks. The package exports a checkpoint-audit template with an explicit unknown-overlap state. Providing a manifest documents a claim; it does not verify it automatically. Freeze the installed package version as well, since an unpinned remote model or interface can change.

Interpret results in three distinct settings. A retrospective pretrained evaluation with uncertain overlap measures transfer performance under unresolved exposure. An overlap-screened historical subset reduces identified risks but remains conditional on the completeness of the audit. A prospective evaluation made with already-frozen weights excludes those weights having seen later test releases, provided all downstream fitting also follows the information contract. A synthetic-only or independently trained checkpoint can be a useful control, but its data provenance must also be inspected.

No retrospective result in this proposal is labeled leakage-free merely because its local features were masked correctly. Until checkpoint evidence is available, a real Chronos run should be described as an exploratory retrospective transfer experiment.

\section{Executed Synthetic Demonstration}
\subsection{Setup and numerical results}
The packaged demonstration generates two monthly series of 132 observation periods beginning in January 2008, with random seed 17. A drifting latent level and a serially correlated initial-publication error generate four archived versions per observation, arriving at 0, 30, 90, and 180 days after first publication. Revision error decays with age. This noise-like mechanism is an engineering fixture, not an economic model or an estimate of ALFRED revision behavior.

Forecasts cover January 2016 through December 2017. There are 48 predictions per method and target: 24 origins for each of two series. Each origin requires at least 24 purged mean-fitting examples and 12 calibration pairs. The targets are first publication and the value in force 180 days later. Table~\ref{tab:demo} contains results produced by the accompanying script. No Chronos weights are used in this experiment.

\begin{table}[t]
\centering
\caption{Executed synthetic demonstration only. Each row has $n=48$. Coverage is for nominal 80\% intervals.}
\label{tab:demo}
\small
\begin{tabular}{@{}llrrr@{}}
\toprule
Method & Target & MAE & CRPS & Cov.\\
\midrule
AR & First & 0.979 & 0.677 & 0.812\\
Revision-linear & First & 1.037 & 0.766 & 0.833\\
State-space & First & 1.033 & 0.722 & 0.896\\
Hindsight AR & First & 1.173 & 0.809 & 0.646\\
\midrule
AR & Mature & 0.748 & 0.534 & 0.875\\
Revision-linear & Mature & 0.717 & 0.548 & 0.875\\
State-space & Mature & 0.795 & 0.564 & 0.854\\
Hindsight AR & Mature & 0.671 & 0.472 & 0.792\\
\bottomrule
\end{tabular}
\end{table}

Revision-linear reduces mature-target MAE by 0.031 relative to AR in this run, but its mature-target CRPS is worse. The paired six-month-block bootstrap interval for the MAE difference is $[-0.148,0.195]$. This example therefore does not support a reliable overall advantage of revision features. It demonstrates why point accuracy and probabilistic performance should be assessed separately.

Hindsight AR has a lower mature MAE than honest AR, with a paired difference of 0.077 and an interval of $[-0.045,0.203]$. For the first-release target, the direction reverses: honest minus hindsight MAE is $-0.194$, with interval $[-0.362,-0.017]$. Later revisions can remove information correlated with early reporting error. Contamination does not have to improve every metric to invalidate the historical information set.

\begin{figure}[t]
\centering
\includegraphics[width=\columnwidth]{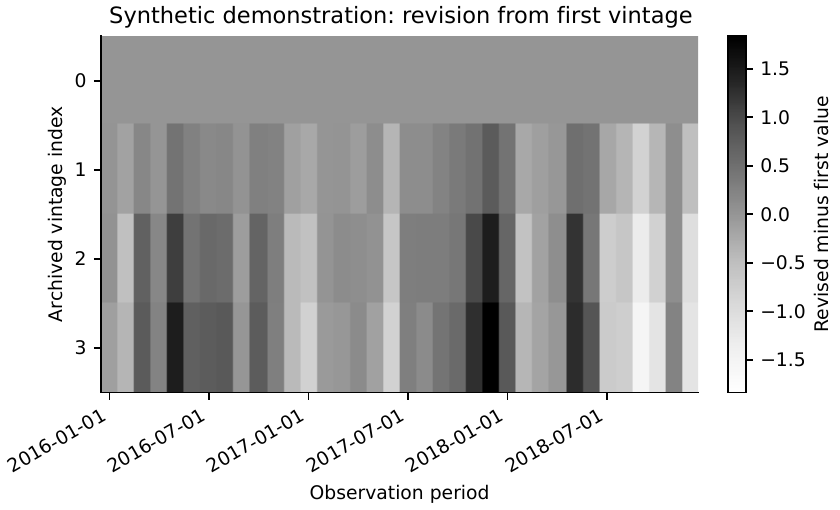}
\caption{Synthetic revision triangle. The color scale records signed change from the first archived value; vintage index counts archived versions, not a universal calendar lag.}
\label{fig:heatmap}
\end{figure}
\begin{figure}[t]
\centering
\includegraphics[width=\columnwidth]{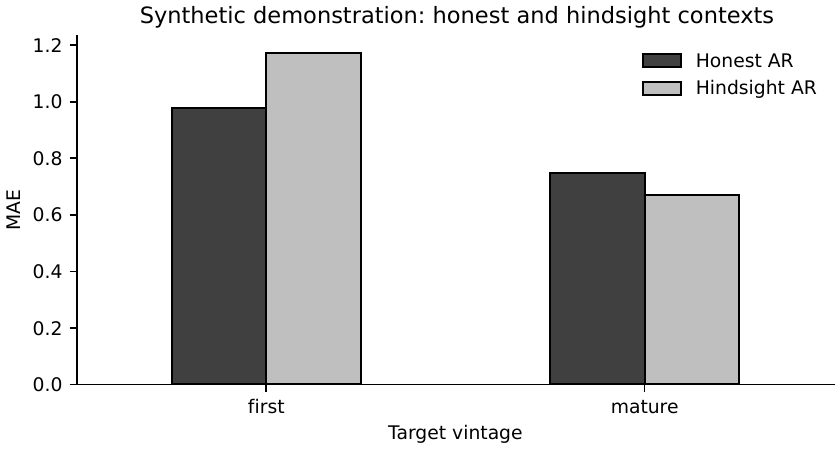}
\caption{The same synthetic issuance dates yield different errors when historical context is replaced by later revisions. The sign of the effect depends on the target vintage.}
\end{figure}
\begin{figure}[t]
\centering
\includegraphics[width=\columnwidth]{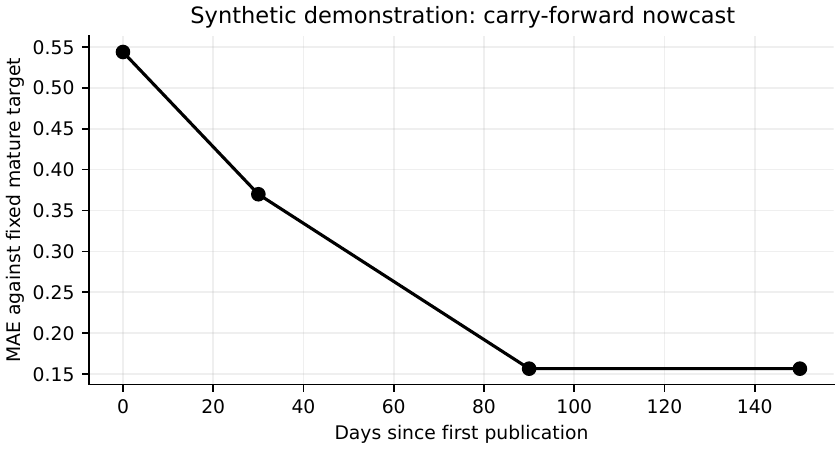}
\caption{Synthetic carry-forward nowcast error against the 180-day target. The decline is expected under the chosen decaying-error generator and is not evidence that all real revisions become easier with age.}
\end{figure}

\subsection{Verification and limits of the artifact}
Thirty-one executed automated tests extend the original nine checks. They cover calendar-date validation, closed and missing validity intervals, right-censored and fixed-maturity targets, causal filling, chronological calibration purging, stable fitting with collinear features, Gaussian-score transformations, and positive joint covariance. An end-to-end test alters every future revision value and verifies that honest forecast means and uncertainties remain unchanged. The information-audit CSV exposes fitting and calibration cutoffs at every origin.

Offline integration fixtures additionally test ALFRED pagination, preservation of missing values, and credential-safe error handling, together with Chronos past-only covariate routing and forecast-date alignment. These are mocked interface tests: the optional Chronos adapter was not executed against downloaded weights, and the live ALFRED downloader was not exercised with an API key. These integration limits are recorded in the package. The artifact is sufficient to reproduce the synthetic results and begin the audited real-data study; it is not evidence that the proposed neural extension or economic benchmark has been completed.

\section{Extended Synthetic Tests}
\label{sec:extended}
\subsection{Design and reproducibility}
To test whether the original demonstration generalizes across random draws and revision settings, we execute a fixed 25-configuration suite. The seeds are 17, 29, 43, 71, and 101. For each seed, we test 90-, 180-, and 365-day mature targets at the standard revision-error scale, plus scales 0.25 and 2 at the 180-day target. Each configuration retains two series, 132 observation periods, and the same 24 evaluation months. All configurations are reported; no hyperparameters are retuned using their outcomes.

This extension uses five archived versions at 0, 30, 90, 180, and 365 days after first publication. It is a separate generator run from the original four-version demonstration. The extra version makes the 365-day target numerically distinct from the 180-day vintage. For revision stress, the multiplier applies to the complete release deviation from the latent path, including residual release noise. Identical seeds retain identical latent paths and innovations across scale settings.

Each configuration produces 384 method/target forecast rows, giving 9,600 rows in total. Those rows reuse observation dates and underlying paths across configurations; they are not 9,600 independent origins. Configuration files are written before evaluation, and every run saves its synthetic fixture, forecasts, information audit, metrics, bootstrap contrasts, and environment manifest. Error bars in the sensitivity plots show one sample standard deviation across five seeds, not confidence intervals.

\begin{figure*}[t]
\centering
\includegraphics[width=\textwidth]{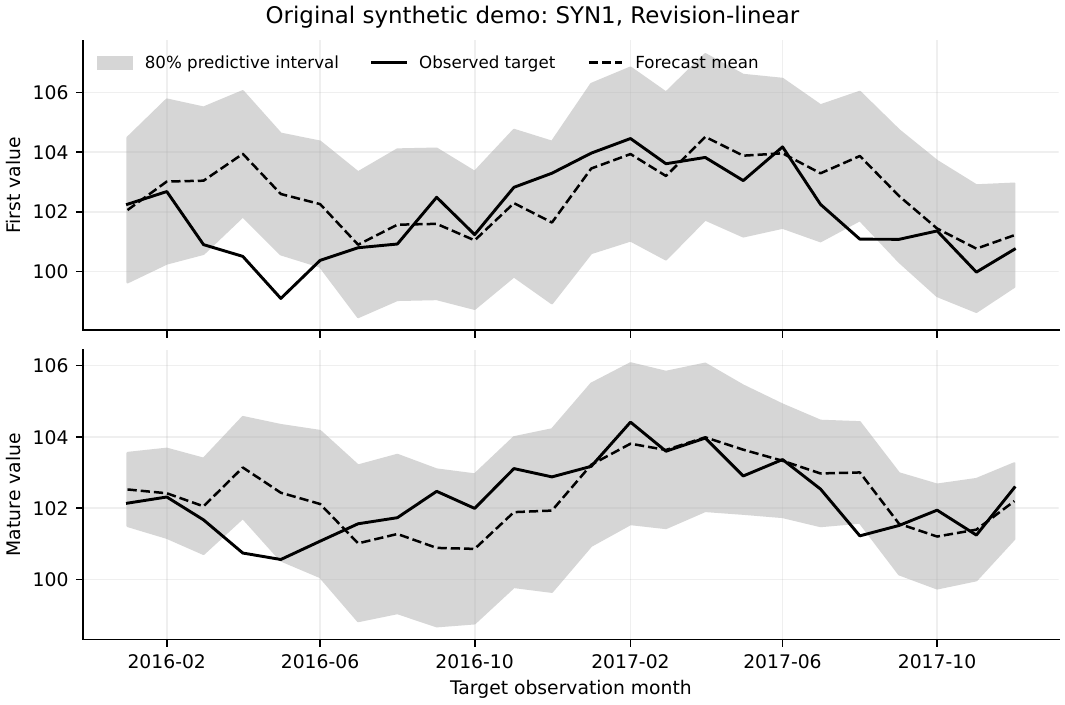}
\caption{original synthetic demonstration, series SYN1. Revision-linear means and 80\% intervals are plotted against first and mature targets over all evaluation months. These are successive rolling forecasts, each issued at its own origin; the connected curves are not one multi-step forecast path.}
\label{fig:forecast_intervals}
\end{figure*}

\begin{figure*}[t]
\centering
\includegraphics[width=\textwidth]{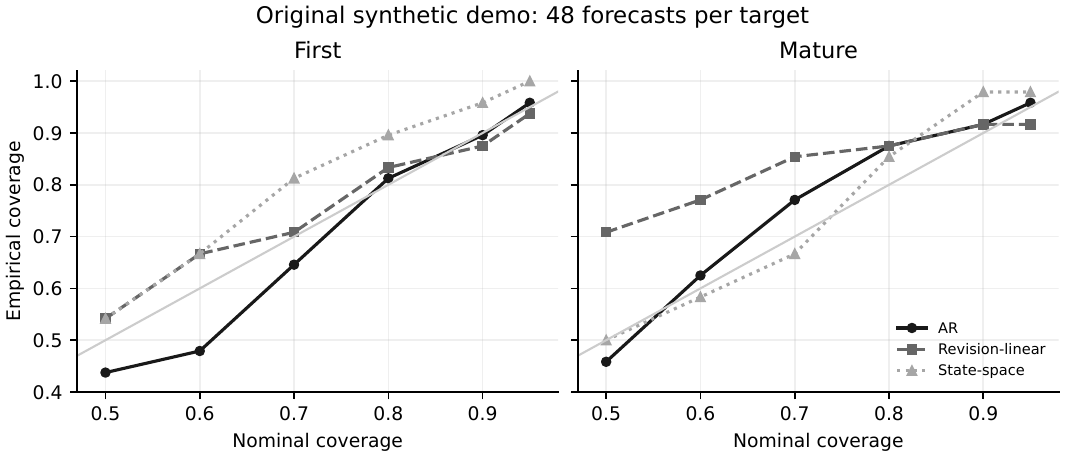}
\caption{nominal versus empirical central-interval coverage in the original demonstration, with 48 forecasts per method and target. The diagonal denotes exact marginal coverage. Departures are descriptive; the temporally dependent observations do not justify independent-binomial confidence statements.}
\label{fig:coverage_reliability}
\end{figure*}

\subsection{Trajectories and calibration}
Figure~\ref{fig:forecast_intervals} complements aggregate errors with rolling predicted means and intervals. Both targets share the same selected series and evaluation dates, avoiding selection of a favorable time segment. The plot can reveal systematic lag, excessive width, and periods where observations fall outside nominal intervals. It does not establish simultaneous path coverage.

Figure~\ref{fig:coverage_reliability} expands the single 80\% coverage statistic to nominal levels of 50\%, 60\%, 70\%, 80\%, 90\%, and 95\%. The same held-out forecast means and scales define every interval. This avoids recalibrating separately against each plotted test coverage. The curves concern marginal Gaussian intervals and remain distinct from the joint first/mature dependence assumption.

\subsection{Variation across seeds}
Table~\ref{tab:extended} summarizes the five-seed, standard-revision, 180-day setting. Revision-linear's mean mature MAE is 1.231, compared with 1.007 for AR and 0.918 for the revision state-space reference. Its mature CRPS is also worse: 0.875 versus 0.728 and 0.656, respectively. Thus, the small mature-MAE gain observed in the original single-seed demonstration does not provide a robust general advantage for this linear adaptation.

\begin{figure*}[t]
\centering
\includegraphics[width=\textwidth]{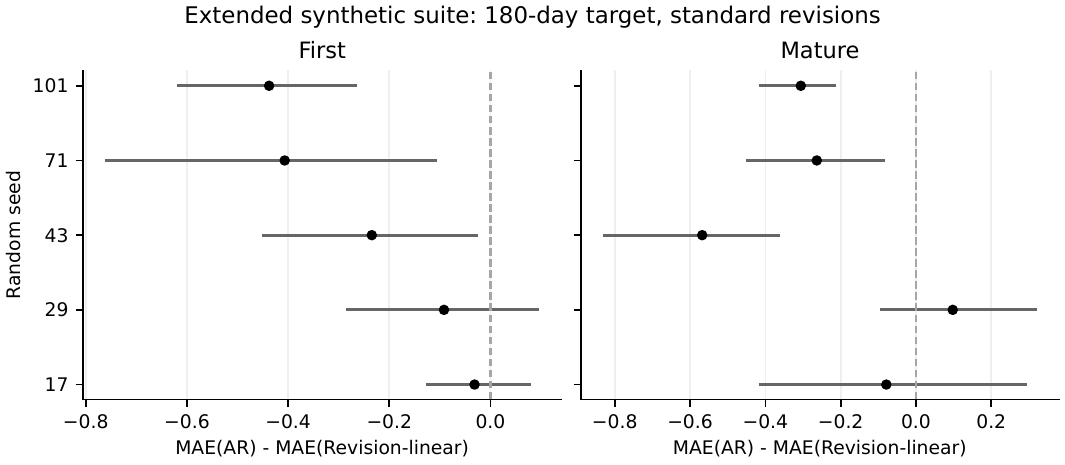}
\caption{per-seed paired MAE differences at 180 days and standard revision scale. Segments are 95\% within-seed percentile intervals from 1,000 six-month circular block-bootstrap draws, with both series retained together within a date. Negative values favor AR.}
\label{fig:seed_robustness}
\end{figure*}

The first-release comparison also disfavors Revision-linear on average. Figure~\ref{fig:seed_robustness} shows paired AR-minus-Revision-linear differences for every seed. Positive differences favor revision features; negative differences favor AR. Three seeds have within-seed 95\% block-bootstrap intervals entirely below zero for both targets. These intervals are descriptive, are not multiplicity-adjusted, and do not substitute for evidence on real releases. No post-hoc model changes were made to reverse the unfavorable result.

\begin{table}[t]
\centering
\caption{Extended synthetic suite at 180 days and standard revision scale. Mean across five seeds; 48 forecasts per seed and target. These differ from the original four-version demonstration.}
\label{tab:extended}
\small
\begin{tabular}{@{}llrrr@{}}
\toprule
Method & Target & MAE & CRPS & Cov.\\
\midrule
AR & First & 1.170 & 0.830 & 0.800\\
Revision-linear & First & 1.411 & 1.001 & 0.750\\
State-space & First & 1.190 & 0.831 & 0.850\\
\midrule
AR & Mature & 1.007 & 0.728 & 0.762\\
Revision-linear & Mature & 1.231 & 0.875 & 0.754\\
State-space & Mature & 0.918 & 0.656 & 0.787\\
\bottomrule
\end{tabular}
\end{table}

\begin{figure*}[t]
\centering
\includegraphics[width=\textwidth]{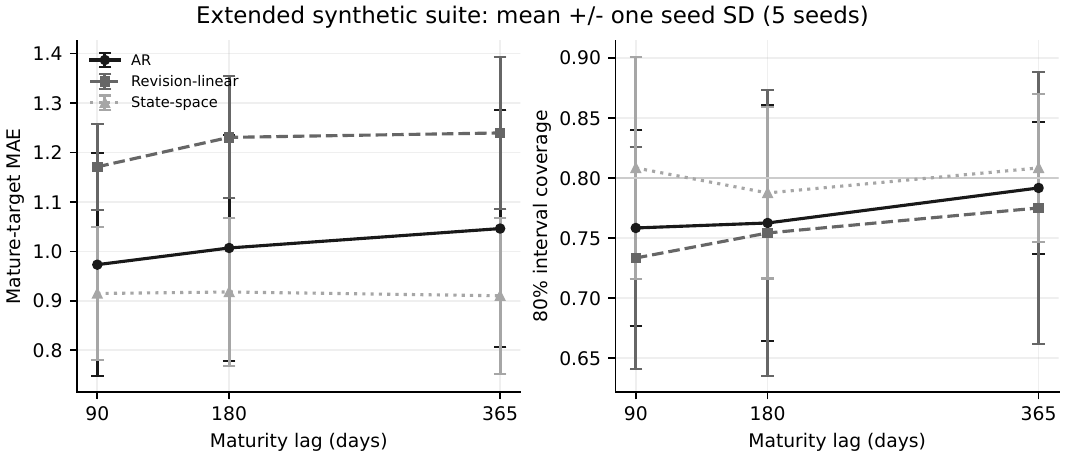}
\caption{mature-target MAE and 80\% coverage versus maturity lag in the extended suite. Points average five seeds; bars show one seed standard deviation. Changing maturity changes both the scored vintage and the availability of training labels, so this is a joint target-and-supervision sensitivity.}
\label{fig:maturity_sensitivity}
\end{figure*}

\begin{figure*}[t]
\centering
\includegraphics[width=\textwidth]{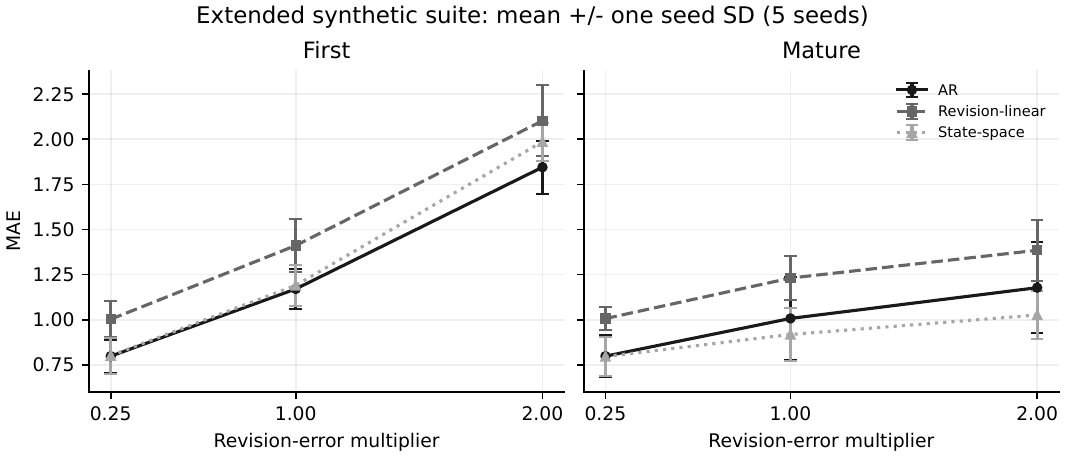}
\caption{first-release and mature-target MAE as release-error amplitude changes. The 180-day target, latent paths, and random innovations are paired within seed. Points average five seeds; bars show one seed standard deviation, not confidence intervals.}
\label{fig:revision_stress}
\end{figure*}

\subsection{Maturity and revision stress}
Figure~\ref{fig:maturity_sensitivity} shows that a later target does not automatically yield a better trained forecaster. AR's mean mature MAE rises from 0.973 at 90 days to 1.046 at 365 days, while the state-space reference remains near 0.91. A longer lag both changes the label and withholds recent supervised examples for longer; the experiment does not identify these effects separately. The state-space reference does not use the same paired-label regression window, which helps explain why the mechanisms differ.

As the revision-error multiplier increases from 0.25 to 2, mature MAE rises from 0.798 to 1.177 for AR, from 1.006 to 1.385 for Revision-linear, and from 0.796 to 1.026 for state-space (Figure~\ref{fig:revision_stress}). This controlled noise-like generator favors an explicitly modeled decaying-error structure; it does not establish state-space dominance under news revisions, structural changes, or an actual economic release process.

Together, these additional tests expose an empirical weakness of the high-dimensional linear revision head with limited supervision. They motivate stronger regularization, better feature compression, and a properly controlled foundation-model comparison as future development work, rather than supporting an unqualified VINTAGE-TS accuracy claim. The real ALFRED and Chronos-2 experiments remain unrun.

\section{Limitations and Study Completion Criteria}
The reference study is deliberately narrow: monthly levels, per-series adaptation, fixed issuance dates, Gaussian joint uncertainty, a small calibration sample, and a summarized revision history. It does not model publication-time uncertainty, arbitrary ragged mixed-frequency panels, benchmark redefinitions, or changing institutional revision policies. Summary revision features can discard path information, and a Gaussian head can miss asymmetry and heavy tails. State-space defaults are not an optimized competing estimator.

The current implementation addresses next-period release forecasts. Mature-value nowcasting of already-published observations is illustrated only by the lag diagnostic and remains a model-training extension. Complete-pair filtering delays the use of first-release supervision. The proposed masked neural loss would alleviate that restriction but requires a tested training implementation and a tuning budget matched to baselines.

Before reporting research conclusions, the study must complete archive-coverage verification, checkpoint provenance analysis, prespecified data transformations, scale-normalized aggregation, adequate real-data rolling evaluation, and the matched Chronos comparison. A convincing outcome requires gains on a declared primary score, paired uncertainty, calibration evidence, and stability across relevant cohorts. A null result is scientifically meaningful: revision histories may be weakly informative, or a latest-vintage model may already exploit their useful consequences.

\section{Conclusion}
\model{} frames revision-aware forecasting as a two-time information problem with explicitly named target vintages. Its central design is to preserve release history, restrict both inputs and labels to what was available at issuance, and distinguish uncertainty about first publications from uncertainty about subsequent changes. A frozen Chronos-2 adaptation offers an implementable starting point, while a joint event-based neural model remains a proposed extension.

The executable package validates the temporal contract with 31 automated tests and supplies a 25-configuration synthetic sensitivity suite. The expanded tests do not establish a consistent benefit of the linear revision head and reinforce the need for strong revision-aware baselines. Its current numerical results establish neither an economic forecasting gain nor a foundation-model improvement. The decisive next evidence is a matched, audited, real-vintage evaluation whose conclusions survive calibration checks and explicit examination of pretraining exposure.

\FloatBarrier
\bibliographystyle{plain}
\bibliography{references}
\end{document}